\documentclass[runningheads]{llncs}
\usepackage{graphicx}
\usepackage[]{algorithm2e}
\usepackage[all]{xy,xypic}
\usepackage{multirow}
\usepackage{array}
\usepackage{cite}
\usepackage{hyperref}
\newcommand\samethanks[1][\value{footnote}]{\footnotemark[#1]}
\makeatletter
\def\thanks#1{\protected@xdef\@thanks{\@thanks
        \protect\footnotetext{#1}}}
\makeatother
 
\begin{document}
\title{Word Sense Disambiguation using  Diffusion Kernel PCA}
%
%
\author{Bilge Sipal\orcidID{0000-0003-0875-1899 } \and
Ozcan Sari\thanks{O. Sari, A. Teke, N. Demirci : \,These authors contributed equally to this work.}\orcidID{0000-0002-9289-0148} \and\\
Asena Teke\samethanks\orcidID{0000-0003-2097-4024} \and
Nurullah Demirci \samethanks\orcidID{0000-0001-7680-7239}}
\authorrunning{B. Sipal et al.}
%
\institute{Istanbul Kultur University, Istanbul, Turkey \\
\email{b.sipal@iku.edu.tr,\\  \{1600003565, 1502010017, 1502010035\}@stu.iku.edu.tr }}

\maketitle              
\begin{abstract}
 One of the major problems in natural language processing (NLP) is the word sense disambiguation (WSD) problem.  It is the task of computationally identifying the right sense of a polysemous word based on its context. Resolving the WSD problem boosts the accuracy of many NLP focused algorithms such as text classification and machine translation. In this paper, we introduce a new supervised algorithm for WSD, that is based on Kernel PCA and Semantic Diffusion Kernel, which is called Diffusion Kernel PCA (DKPCA). DKPCA grasps the semantic similarities within terms, and it is based on PCA. These properties enable us to perform feature extraction and dimension reduction guided by semantic similarities and within the algorithm. Our empirical results on SensEval data demonstrate that DKPCA achieves higher or very close accuracy results compared to SVM and KPCA with various well-known kernels when the labeled data ratio is meager. Considering the scarcity of labeled data, whereas large quantities of unlabeled textual data are easily accessible, these are highly encouraging first results to develop DKPCA further. 

\end{abstract}

\section{Introduction}

Nearly $\%40$  words in the English language have estimated to have ambiguous meanings, i.e., their meanings depend on their contexts.  There are two documented reasons for such an ambiguity; Homonymy which means that an ambiguous word can have the same spelling or pronunciation with a word that has a different sense or polysemy which means that the ambiguous word can have more than one sense \cite{Nguyen13}. 
For example, the word \textit{interest} can have six different meanings (see Table \ref{int}).  If it is used as in \textit{interest in this business}, then the meaning is to give attention (see sense 3 in Table \ref{int}), and the usage \textit {ten percent interest in this business} has another meaning which means a share in a company   (see Sense 5 in Table \ref{int}).  There is a clear semantic link between these two sentences whereas the homonymous words are semantically unrelated such as the usage of the word \textit{bank} as in \textit{the savings bank} and the usage as in \textit{a river bank} \cite{McCaughren09}. 
Hence the  main problem is to predict the exact sense of the word, i.e., choosing the right class defined by the dictionary based on its context. The generalized name for resolving such problems is Word Sense Disambiguation (WSD) which is to computationally identify the exact sense of an ambiguous word from its context \cite{Navigli09}. Compared to polysemy, the homonymy is more natural to resolve since the context of the words are in general very different. Polysemy, on the other hand, can be seen in very similar contexts as in the example, and it may become a very subtle problem. 
 Native speakers tend to understand these subtle changes in the meaning subconsciously, and catching the exact meaning seems effortless to them.  On the other hand, for  computers as well as language learners, solving the problem is a nontrivial task  \cite{Turdakov10}.   
  
  WSD  is handled extensively as a part of text classification studies since it potentially has significant effects on choosing the right class of a given document.  Moreover, it is used as a part of machine translation algorithms \cite{Carpuat07}. There are two kinds of methods that are used in attacking a  WSD problem, one of them is called knowledge-based, and the other one is called the corpus-based method \cite{Navigli09}. Knowledge-based methods take advantage of the knowledge resources such as dictionaries whereas corpus-based methods use manually sense annotated datasets to train a model.  For machine learning, the general approach is to use corpus-based methods since the performance of supervised methods are higher than the unsupervised ones \cite{Li16}.  
  Another approach for resolving the WSD problem is to use Kernel methods  \cite{Cortes95, Muller01, Shawe-Taylor04}. The basic idea of Kernel methods is to get the non-linear similarities of the data without explicitly computing the feature maps via kernel trick \cite{Scholkopf02}. The problem of choosing the right kernel for the right task is called model selection \cite{Gonen11}.  General usage favors Support Vector Machines (SVM) as the classifier \cite{Cortes95, Muller01, Shawe-Taylor04,Wang14, Wang17, Wang14}.  On the other hand, there are kernel studies that make use of  Principal Component Analysis (PCA), which is called Kernel Principal Component Analysis (KPCA) \cite{ Scholkopf98, Su04,Wu04}.  For further reading on kernel methods, we refer to \cite{Hofman08}, and for further knowledge of the  usage of kernels for WSD, we refer to \cite{Li16}.

In this study, we introduce a corpus-based kernel method, which is called 'Diffusion Kernel Principal Component Analysis' (DKPCA). Following the footprints of \cite{ Kandola03, Wang14, Wu04}, we merge the semantic diffusion kernel and non-linear PCA in order to construct DKPCA.  In section  \ref{SemKer}, we explain Bag of Words (BoW) representation and its shortcomings for grasping intrinsic semantic relations of terms. Such drawbacks of this representation lead to the demand for semantic similarity kernels, and we show the construction of such kernel, i.e., semantic diffusion kernel.  We wrap up the mathematics and give an algorithm to compute semantic diffusion kernel.   In section \ref{KPCA}, we give a linear algebra based approach to the well known KPCA algorithm \cite{Vidal16}. By using the results from section \ref{KPCA}, we introduce DKPCA. 

We test our algorithm with SensEval data and compare it with SVM  and KPCA algorithms with several kernels. For evaluating our algorithm, we use mainly $F_1$ scores, since the label distribution of our datasets is imbalanced. Our $F_1$ macro-averaged  results  indicate that DKPCA outperforms all the competitive algorithms for $5\%$ and $10\%$ labeled data in SenSeval task (interest, serve, line, and hard) and get similar results for $\%30$ labeled data. For   $F_1$ micro averaged scores, we  get close results with SVM with semantic diffusion kernel for hard data set  and DKPCA still outperforms  the other algorithms $5\%$ and $10\%$ labeled data for interest, serve and  line  datasets. These results are highly encouraging concerning the scarcity of the labeled data.

\section{Semantic Diffusion Kernel for WSD}

\subsection{Problem Set up for WSD}\label{SemKer}
Let $\{t_0, t_1,\dots, t_N \}$ denote the terms (words) in our dictionary which is the set of terms appear in the corpus (the set of documents or contexts) and let $\{d_1,...,d_m\}$ denote the set of documents.  Let $t_0$ be disambiguated which is seen in the document $d.$ We remove $t_0$ from the document $d$ and define the following map
\begin{eqnarray*}
\varphi: \mathcal{D} &\longrightarrow& \mathcal{D}\\
d_i  &\mapsto &(tf(t_1,d_i)...tf(t_N,d_i))=x_i
\end{eqnarray*}where $tf(t_j, d_i)$ represents the frequency of the term $t_j$ in document $d_i.$ Then the $document \times term$  data matrix $D$ is given as follows. 
\begin{equation}
\label{datD}
D= \bordermatrix{& t_1 & \dots &t_N \cr
              d_1 & tf(t_1, d_1) &\dots & tf(t_N, d_1) \cr
                    \vdots   &  \vdots       & \ddots& \vdots\cr
               d_m&   tf (t_1, d_m)   & \dots & tf(t_N, d_m)  \cr}
               \end{equation} This representation of the   data is called the Bag of Words (BoW) representation. The matrix  
\begin{equation}
\label{linker}
K=DD^T
\end{equation}
is the Gram matrix that represents the following inner product
\begin{eqnarray}
K:  X\times X&\longrightarrow&\mathrm{R}^N\\
 \,    \, \,   (x_i,x_j)&\mapsto &(x_i,x_j^T)\nonumber
 \label{innermap}
\end{eqnarray} where $D^T$ is a $term \times document$ matrix \cite{Shawe-Taylor04}, and $K$ is a $document \times document$ matrix. 
Note that we use the same symbol $K$  for the Gram matrix and the inner product map since they are equivalent.
The Kernel Map (\ref{innermap}) and the matrix  Kernel  (\ref{linker}) defined above both indicate  the most basic kernel, namely the linear kernel, that satisfies the Mercer Conditions. These conditions  force the given matrix to be positive semi-definite and symmetric in order to be a valid Kernel function. More sophisticated functions can be chosen as a kernel candidate as long as the representation matrix satisfies the Mercer conditions. Some popular kernels are as follows.
\begin{eqnarray}
\label{kernels}
\text{Gaussian \, Kernel}&:& K_{rbf}(x_i, x_j)=e^{\frac{-||x_i-x_j||}{2\sigma^2}}\label{ilk}\\
\text{Polynomial \, Kernel}&:& K_{pol}(x_i, x_j)= (x_ix_j +1)^d \label{iki}\\
\text{Linear \, Kernel}&:& K_{lin}(x_i, x_j)=x_i ^T x_j \label{uc}
\end{eqnarray}

\subsection{Semantic Diffusion Kernel}\label{SemKer1}
We can represent a data matrix by employing an undirected graph where each vertex indexes each term and each  edge keeps the information of the co-occurrences between terms in the documents. 
Let us explain what we mean by 'co-occurrences between terms' with an example.   Let us consider three  sentences:   \textit{Today is very cold and dark} is the first one, \textit{Dark rooms have generally have mold} is the second one and \textit{Mold can cause sickness}  is the third one. We call them as $d_1, d_2,$ and $d_3$ respectively.  We focus on the terms \textit{cold}, \textit{dark}, \textit{mold} and \textit{sickness}, which we name $t_1, t_2, t_3,$ and $t_4$ respectively.   The terms  $t_1$ and $ t_2$ have a first order co-occurrence since they are both in $d_1.$   Similarly, terms   $t_2$ and $t_3$  with terms  $t_3$ and $t_4$ have first order co-occurrence. This way, we can construct the representation matrix $D^T$   as given in \cite{Wang14, Wang17}. 
\begin{equation}
(D^{T})_{ij} = \left\{
\begin{array}{rl}
1 & \, \, \text{if   } \, \,t_j \in d_i,\\
0 &\, \, \text{if    } \, \,otherwise.
\end{array} \right.
\end{equation}
Wang et al. \cite{Wang14, Wang17} show that Matrix $G=D^TD$ captures the first order correlations since it gives out the first order co-occurrence paths between terms.
However, the first order co-occurrences indicate a substantial similarity between the terms, the representations that depend on them like  BoW representation and the graph representation, are both sparse and only carries limited information. For example,  there is a semantic similarity between \textit{cold} and \textit{mold}, which is maybe crucial in classifying $d_1$ and $d_2$, but $G$  fails to represent this phenomenon.

Hence, careful observation will reveal that the similarity information of semantically close yet distinct words is lost in these representations.  A common approach to solve this problem is to enrich BoW representation with the usage of semantic kernels. A semantic kernel  $S$ is a $term\times term$ matrix which carries the global similarity scores among terms. It is used to diffuse this information to the kernel in use without harming the Mercer conditions (positive definiteness and the symmetry). 
In Equation (\ref{semdif}) we diffuse a semantic kernel $S$ to the   linear (BoW) kernel $K_{lin}.$
\begin{equation}
\label{semdif}
K_{new}(x_i, x_j)=x_i  S S^Tx_j 
\end{equation}
There are different ways to create a semantic kernel \cite{Altinel15, Cristianini02, Gliozzo05} but how can we construct a semantic kernel $S$ that can be used as a vessel to carry this semantic information? Before answering this question, let us go back to our example. If there is a first order co-occurrence between two terms, then they are said to have first order correlation, e.g., \textit{cold} and \textit{dark} have a first order correlation.    The similarity between the term \textit{mold} and the term \textit{cold} is called the second order correlation through the term \textit{dark}. Moreover, there is a third order correlation between \textit{cold} and \textit{sickness} through cold and mold. Such correlations among terms are called higher order correlations and can be captured by $G^p$ where $p$ is the order. The answer to our question is the semantic diffusion kernel in Equation (\ref{Tay})\cite{Kandola03}, which  is a matrix that is  designed to carry out the high order semantic correlations.
\begin{eqnarray}
\label{Tay}
S&=& e^{\lambda G/2}\\
&=& (G^0+ \lambda G + \frac{\lambda^2 G^2}{2!} +\dots +\frac{\lambda^p G^p}{p!}+\dots )\nonumber\\ 
&=&\sum_{p=0}^{\infty} \frac{\lambda^p G^p}{p!}\nonumber
\end{eqnarray}
 Intuitively, as the order of the correlation increases the similarity score should decrease. That is why we use  $\lambda$  as the decaying factor in Equation \ref{Tay}, which is the Taylor expansion of $e^{\lambda G/2}.$ As the sum goes to infinity, this kernel approaches the Gaussian kernel. In essence, the diffusion kernel is the discretized Gaussian kernel, which is the solution of the time-dependent (continuous) diffusion equation \cite{Kondor04}.  In practice, the semantic problem does not need a continuous construction at all. Our experiments show that even after the second or third step of the Taylor expansion, the similarity scores among terms fade away.  
 Now let us check whether Mercer conditions hold.  The matrix $S$ is a $term \times term$ symmetric, positive semi-definite matrix. It encodes the semantic relations among terms (features). This semantic kernel has to be diffused to the linear Kernel (\ref{linker}) as follows. 
 \begin{equation}
 K_{sd}=DS(DS)^T=DSS^TD^T
 \end{equation} Matrix $K_{sd}$ is a symmetric and positive semi-definite. 
 Let us sum up the discussion above in the following algorithm.\RestyleAlgo{boxruled}
\LinesNumbered
\begin{algorithm}[ht]
  \caption{Diffusion Kernel Algorithm\label{algDifKer}}
 \KwData{ Matrix $D$, Parameter $\lambda,$ Step}
 \KwResult{Semantic Diffusion Kernel K}
 \textbf{initialization}\;
 Calculate $G=D^TD$\;
 Compute the Taylor expansion as follows\;
 $S:$ Identity matrix of the same size as $G$\;
  \For {$I =1\, \textbf{to} \,  \, Step$}{
   $S:= S+\frac{ \lambda ^I} { I!} G^{I}$\;
  }
  Compute $K_{sd}=DSS^TD^T$\;
\end{algorithm}

 \section{Kernel Principal Component Analysis}\label{KPCA}
Principal Component Analysis (PCA) is a technique that captures the information from data by a linear transformation to a coordinate system which is mathematically designed to maximize the variance among data points \cite{Pearson01, Hotelling33}. If the data does not fit in a linear subspace of the space that data lies then PCA fails to perform this task.  A general method to deal with such non-linearity is Kernel PCA (KPCA) \cite{Scholkopf98}. A Kernel function or a  Kernel matrix is calculated directly from the data and represents a dot product in a feature space.  It enables us to get the non-linear similarities of the data without explicitly computing the feature maps via kernel trick \cite{Hofman08}.
Mathematically, KPCA is a method to compute the non-linear principal components, which correspond to the eigenvector of the Kernel matrix \cite{Vidal16}.   
    If we focus on the Kernel methods what we observe that the basic idea of such methods is to map the data points from the input space to some feature space where the separation of the data becomes computationally more manageable, this simplicity stems  from the so-called 'Kernel Trick.' Kernel trick enables us to skip the cumbersome mathematics of computing the images of $D$ in a feature space and projecting back the result of the inner product in that space. Instead, we can construct an inner product with specific criteria, and we have all the similarities of the data in feature space.

 As discussed above,  KPCA  projects the data a non -linear coordinate system which describes the intrinsic relations among data points better than a linear coordinate system.  The first step to achieve this goal is to construct the right kernel matrix. In this work, to capture the semantic similarities, best possible kernel among custom kernels is the semantic diffusion kernel. 
 
For the problem setup,  we follow  the notation of  \cite{Vidal16} and we refer this excellent source for detailed information on the topic.   We construct the semantic diffusion kernel $K_{sd}$ by following Algorithm (\ref{algDifKer}). To get a meaningful variance, we center the kernel matrix $K_{sd}$ as 
\begin{equation}
\label{centerJ}
\hat{K}_{sd}= JK_{sd} J^T \, \, \text{where} \, \, J_{ij} = 
\left\{
\begin{array}{rl}
 1 & \, \, \text{if   } \, \,i=j \\
   \frac{-1}{m}&\, \,   \, \,otherwise.
\end{array} \right.
\end{equation}Matrix $J$ in Equation (\ref{centerJ}) is called the centering matrix, and it is an $M\times M$ matrix which has $1$ on the diagonal and $\frac{-1}{m}$ on the other positions. After centering we compute the eigenvalues and the corresponding eigenvectors of $\hat{K}_{sd}.$ 

\begin{equation}
\label{eig}
\hat{K}_{sd} w_i= \lambda_i  w_i
\end{equation}
The eigenvalues are $\lambda_i,$ and $w_i$ are the corresponding eigenvectors. We let $\Lambda$ denote the diagonal matrix that contains the eigenvalues and $W$ denote the matrix of eigenvectors.  Then Equation (\ref{eig}) becomes
 \begin{equation}
\label{EV}
\hat{K}_{sd} W= \Lambda W.
\end{equation}
However, our main purpose is to find the right transformation which enables us to separate the data as accurate as possible,  before the projection step we can also apply the dimension reduction as in traditional PCA  in order to clear the noise. This step is crucial when the number of features is very high. Let us  choose the dimension  $d$ and take the first $d$ eigenvalues $\Lambda_d$ and corresponding eigenvectors $W_d.$ For choosing this number either grid search on the ratios of the eigenvalues or by using a graph can be applied.

After ordering the eigenvalues in descending order, we pick the first $d$ eigenvalues and the corresponding eigenvectors. Then we normalize the eigenvectors  so that for every $i\in\{1,..., d\}$ we have  $$||w_i||=\lambda^{-1/2} \iff \hat{W}_d=\Lambda_d^{-1/2} W_d.$$
Then the low  dimensional non-linear principal components  for the entire data set can be calculated as follows.
\begin{eqnarray}
\label{NPC}
Y&=& \hat{W}_d^T K \nonumber \\
&=& \Lambda_d^{-1/2}W^T K\nonumber\\
&=& \Lambda_d^{-1/2}\Lambda W_d^T\nonumber\\
&=& \Lambda_d^{1/2} W_d^T  
\end{eqnarray}
Equation \ref{NPC} implies that the low dimensional coordinates can be computed by using the eigenvalues and eigenvectors of the kernel matrix $K_{sd}.$

\subsection{Approach-Diffusion Kernel PCA Algorithm}\label{DKPCA}
The main goal of this work is to create a methodology which learns  the semantic similarities efficiently and transforms the data while applying dimension reduction guided by these similarities and classify  this processed data.   In the previous sections, we have established  the basic theoretical  tools necessary to construct such an algorithm.  Now we combine these tools to create our algorithm the Diffusion Kernel Principal Component Analysis. 

 We start by preprocessing the data by cleaning punctuation and stop words with NLTK\footnote{https://www.nltk.org/ } and continue to get the BoW representation of it as given in Section \ref{SemKer}. Then we compute the Diffusion Kernel $K_{sd}$ as explained in Section \ref{SemKer1}.  This kernel is a $document \times document$ matrix.  We use this matrix as we use the covariance matrix in plain PCA, i.e., we centralize $K_{sd}$ as in Equation (\ref{centerJ})  and compute the eigenvalues (\ref{eig}) and eigenvectors as in (\ref{EV}). 
 
 At this point, we decide whether a dimension reduction should be applied. The main idea is the find the values that explain the most significant portion of the variance. First, we order the eigenvalues in descending order, and  we did grid search in one fold of the ten folds train-test splits (this part is explained in detail in Section \ref{exp}) to find the most appropriate dimension and avoid data leakage. Our search process checks the ratios of the eigenvalues.  As well checking the ratios of eigenvalues, suitable  dimension can be estimated by checking the ratio of each eigenvalue to the sum of the eigenvalues  and  getting rid of the smallest values.   When all features are crucial for the model, this step can be skipped, but we believe for long texts with a large vocabulary, this step is a vital ingredient to get the right dimension reduction that is guided by the semantic similarities.  Then by applying Equality (\ref{NPC}) we get the non-linear transformation of the data.
 
 These steps were done in an unsupervised manner except for the search for the right dimension.  We need a classifier to convert the algorithm to a supervised algorithm. We choose the  k-nearest neighbors algorithm because of three main reasons.
\begin{itemize}
\item[1)] We want to compare our results with \cite{Wu04} which introduces the usage of (polynomial kernel) KPCA with KNN for WSD and \cite{Wang14} which presents the usage of Diffusion Kernel with SVM
\item[2)] As a result of  Reason (1), SVM as a classifier is not an option  
\item[3)] KNN is an algorithm with high variance and low bias. Choosing KNN disables the classifier to infuse its own bias to our model  and enables us to test the performance of DKPCA better.
\end{itemize}Let us  wrap up the mathematical computation  and our discussion with the following algorithm.
As explained above for the last step,  any standard classification algorithm can be used.

\RestyleAlgo{boxruled}
\LinesNumbered
\begin{algorithm}[ht]
  \caption{Diffused Kernel PCA\label{alg}}
 \KwData{Matrix $X$, Parameter $\lambda,$ Step}
 \KwResult{Non-linear Principal Components Y}
 \textbf{initialization}\;
 Calculate Matrix $D$ as given in (\ref{datD})\;
 Call Algorithm (\ref{algDifKer})\;
 Compute Matrix $J$ as in (\ref{centerJ})\;
Center $K$ as $\hat{K}_{sd} =JK_{sd}J^{-1}$\;
Calculate Matrices $\Lambda$ and $W$ as given in (\ref{EV})\;
Apply dimension reduction from $N$ to $d$ and get $\Lambda_d$ and $W_d$\;
Compute the transformed dataset $Y$ as in (\ref{NPC})\;
Compute KNN with input  $Y$\;
\end{algorithm}
\vspace{-0.5cm}

 \section{Experiments, Evaluations  and Results}\label{exp}
 In order to evaluate the performance of DKPCA for the WSD task, we use the most commonly used  datasets of the SensEval.\footnote{http://senseval.org/} We implemented  SVM and non-linear PCA based kernel methods with linear, polynomial, Gaussian and semantic diffusion kernels.   

 \subsection{Datasets}

 \begin{table}[h!]
\centering
\caption{Senses and Frequencies in Hard Dataset}
\label{hard}
\begin{tabular}{|l|l|l|}
\hline
\bf{No} & \bf{Sense}                                      & \bf{Frequency}  \\ \hline
1  & not easy (difficult)  & 3455      \\ \hline
2  & not soft (metaphoric) & 502       \\ \hline
3  & not soft (physical)   & 376       \\ \hline
\end{tabular}
\end{table}

 \paragraph{Hard Data}  consists of 4333 instances with three senses taken from WordNet, which is given in Table \ref{hard}. The instances are taken from  the San Jose Mercury News Corpus (SJMN) \cite{Leacock98}.
As seen in Table \ref{hard} the data is not equally distributed, nearly three fourth of the data is labeled with sense 1,  which is why  we included $F_1$ micro and macro scores.
\begin{table}[h!]
\centering
\caption{Senses and Frequencies in Line Dataset}
\label{line}
\begin{tabular}{|l|l|l|}
\hline
\bf{No} & \bf{Sense}                                      & \bf{Frequency}  \\ \hline
1  & Stand in line                & 349       \\ \hline
2  & A nylon line                 & 502       \\ \hline
3  & A line between good and evil & 374       \\ \hline
4  & A line from Shakespeare      & 404       \\ \hline
5  & The line went dead           & 429       \\ \hline
6  & A new line of workstations   & 2217      \\ \hline
\end{tabular}
\end{table}\paragraph{Line Data}  is due to \cite{Leacock93}. It consists of 4147 instances which are gathered from  1987-1989 Wall Street Journal(WSJ) corpus and the American Printing House for the Blind  (APHB) with 
six different senses annotated from WordNet.  Similar to hard data, the distribution of the instances are not homogenous. 
\begin{table}[]
\centering
\caption{Senses and Frequencies in Serve Dataset}
\label{serve}
\begin{tabular}{|l|l|l|}
\hline
\bf{No} & \bf{Sense}                                      & \bf{Frequency}  \\ \hline
1  & Function as something & 853       \\ \hline
2  & Provide a service     & 439       \\ \hline
3  & Supply with food      & 1814      \\ \hline
4  & Hold an office        & 1272      \\ \hline
\end{tabular}
\end{table}    \paragraph{Serve Data}  is due to \cite{Leacock98} like hard data and like line data, the data is  gathered from  1987-1989 Wall Street Journal (WSJ) corpus and the American Printing House for the Blind  (APHB). It is labeled with four distinct senses from WordNet.      
\begin{table}[h!]
\centering
\caption{Senses and Frequencies in Interest Dataset}
\label{int}
\begin{tabular}{|l|l|l|}
\hline
\bf{No} & \bf{Sense}                                      & \bf{Frequency} \\ \hline
1  & Readiness to give attention                & 361       \\ \hline
2  & Quality of causing attention to be given   & 11        \\ \hline
3  & Activity, etc. That one gives attention to & 66        \\ \hline
4  & Advantage, advancement or favor            & 177       \\ \hline
5  & A share in a company or business           & 500       \\ \hline
6  & Money paid for the use of money            & 1252      \\ \hline
\end{tabular}
\end{table}\paragraph{Interest Data}   is created with 2368 instances from   part of speech tagged subset of the Penn Treebank Wall Street Journal Corpus (ACL/DCI version) which are annotated by using the Longman Dictionary of Contemporary English \cite{Bruce94}.

\subsection{Experimental Setup}\label{Ex}

In this study, we compare the performances of six different algorithms. Three of them are SVM based: SVM with linear, Gaussian, or radial basis function (RBF) and semantic diffusion kernel.  The rest is PCA based: PCA, Kernel PCA with Gaussian (RBF), Polynomial, and Diffusion Kernel PCA (DKPCA).  All proposed algorithms are implemented with Python\footnote{https:// www.python.org/downloads/} using Python Data Analysis Library (pandas v0.22.0), NLTK, Machine Learning Libraries (scikit learn 0.19.1 and sklearn). For the kernels (RBF, Lin and Pol) and algorithms (PCA, KPCA and SVM) we used the embedded algorithms in  scikitlearn. We implemented Diffusion kernel and Diffusion Kernel PCA by following Algorithms  (\ref{algDifKer})  and (\ref{alg}). Our implementation can  be reached in GitHub\footnote{Github.com/dkpca/wsd-project}. 
All the tests are performed on a computer with CPU:  INTEL(R) CORE(TM)I7-8700CPU@3.20GHZ and 16 GB RAM.

For preprocessing, the stop words with non-alphabetical words are removed and we compute the vector  representation of the data by using tf (term frequency) as in Section \ref{SemKer}.  We split the datasets into three different training set sizes with label percentage: $\%5, \%10$ and $\%30.$  We run the algorithms for ten times with these   train-test  splits.  After testing  for each fold we take the arithmetic mean of  these results as the final evaluation result in Tables \ref{Acc}-\ref{F1-mac}.

\begin{figure}
\centering
\includegraphics[width=7cm]{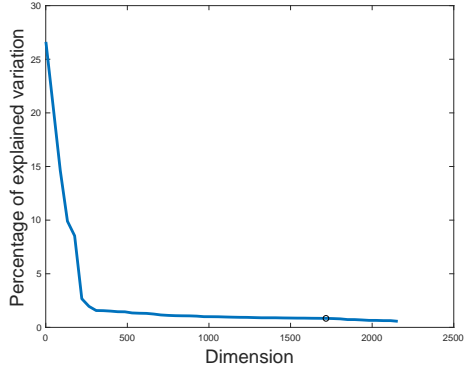}
\caption{For interest data set,  we reduced the  dimension from  6929 to 1710. } \label{fig1}
\end{figure}

In order to avoid data leakage, we optimize parameters only in one fold of these train-test splits.  For the linear kernel, the only optimization is done for parameter $C.$ After grid search for the values $\{0.01, 0.05,  0.1,0.5, 1, 10 \}$ we find out that the values $1, 0.01,0.05, 10$ cause slight increases in the accuracy results depending on the data set. For the RBF kernel, we do not optimize the parameter $\gamma$ since our previous attempts show that the default value in sklearn gives the best accuracy values.  
The choice of the decaying factor $\lambda$ is done by grid search as well and  is fixed as  $\{0.0039\}.$  We are surprised to observe that choice of the Taylor step should be done either 2 or 3 steps. This  makes sense once we consider the effect of the semantic similarity, which is expected to be getting smaller as the correlation is of higher order \cite{Ganiz06}.  
For the polynomial kernel, we use the default degree, which is 3. 
We  perform dimension reduction on the interest dataset by checking the ratios of the eigenvalues. Estimating the right dimension can be done by  calculating  the ratios  of each eigenvalue to the sum of eigenvalues as well.  Interest dataset has  2367 documents (contexts)  and 6929  terms (features). We reduced the  dimension from  6929 to 1710 as seen in Fig \ref{fig1}.  In Table \ref{IntDim}, we present the test results without the dimension reduction and compare the performance in Section \ref{Eva}.
Our main algorithm is based on KPCA and in order to use the unsupervised KPCA as a supervised algorithm, we need to use a classifier other than SVM.  We choose KNN as we explain in detail  in Section \ref{DKPCA}.   Choice  of the number of neighbors  for KNN is done by employing cross validation on the values $\{1,..10\}$ and tests indicate that  it should be 6.

\subsection{Evaluation}\label{Eva}
The average  test results after ten train-test splits are summarized in Tables \ref{Acc}-\ref{F1-mac}.  The abbreviations on the tables are Lin. indicates linear kernel. RBF corresponds to Gaussian kernel. Dif. is the SVM with Diffusion kernel and Pol. is the polynomial kernel. The datasets we use have  skewed label distribution as  given in Tables  \ref{hard}-\ref{int}. As a result, we employ $F_1$ micro and macro scores as our  primary evaluation metric.
\begin{table}[!htp]
\centering
\caption{Accuracy scores on the Datasets: If the difference between two competitive scores is less than $1\%$ then we emphasized both scores.}
\label{Acc}
\begin{tabular}{|l|l|l|l|l|l|l|l|l|}
\hline
Dataset                   & Size \% & SVM Lin        & SVM RBF & SVM Dif.       & PCA   & KPCA Pol. & KPCA RBF & DKPCA          \\ \hline
\multirow{3}{*}{Interest} & 5\%     & 57,36          & 53,00   & 63,86          & 64,17 & 64,37     & 64,32    & \textbf{72,94} \\ \cline{2-9} 
                          & 10\%    & 65,06          & 53,15   & 65,39          & 67,41 & 66,77     & 67,65    & \textbf{77,53} \\ \cline{2-9} 
                          & 30\%    & 79,48          & 53,59   & 79,04          & 73,08 & 72,99     & 72,80    & \textbf{81,68} \\ \hline
\multirow{3}{*}{Line}     & 5\%     & 53,74          & 53,46   & 58             & 60,45 & 62,76     & 62       & \textbf{67,92} \\ \cline{2-9} 
                          & 10\%    & 56,06          & 53,45   & 61,26          & 63,88 & 66,48     & 65,57    & \textbf{71,27} \\ \cline{2-9} 
                          & 30\%    & 73,27          & 53,46   & \textbf{75,51} & 69,21 & 70,58     & 71,71    & \textbf{75,75} \\ \hline
\multirow{3}{*}{Serve}    & 5\%     & 65,27          & 41,44   & 68,71          & 63,53 & 64,04     & 62,06    & \textbf{71,73} \\ \cline{2-9} 
                          & 10\%    & 72,26          & 41,45   & 70,69          & 66,56 & 66,94     & 67,05    & \textbf{74,48} \\ \cline{2-9} 
                          & 30\%    & \textbf{81,71} & 41,48   & 78,51          & 69,81 & 70,88     & 70,47    & 77,23          \\ \hline
\multirow{3}{*}{Hard}     & 5\%     & 79,80          & 79,76   & \textbf{81,81} & 80,19 & 79,76     & 79,80    & \textbf{81,15} \\ \cline{2-9} 
                          & 10\%    & 80,12          & 79,76   & \textbf{82,54} & 80,41 & 79,86     & 79,79    & \textbf{82,31} \\ \cline{2-9} 
                          & 30\%    & 81,79          & 79,85   & \textbf{84,07} & 80,86 & 80,08     & 80       & \textbf{83,86} \\ \hline
\end{tabular}
\end{table}
Tables \ref{Acc}-\ref{F1-mac}  show the results after the dimension reduction and Tables \ref{IntDim} and \ref{NoDim} show the test performance of DKPCA without  dimension reduction. \begin{table}[!htp]
\centering
\caption{DKPCA Scores Without Dimension Reduction for Interest Dataset}
\label{IntDim}
\begin{tabular}{|l|l|l|l|}
\hline
Size & $F_1$-macro & $F_1$-micro & Accuracy \\ \hline
5\%  & 36,38      & 65,51      & 65,51    \\ \hline
10\% & 41,96      & 74,74      & 68,46    \\ \hline
30\% & 50,68      & 74,74      & 74,74    \\ \hline
\end{tabular}
\end{table}\vspace{-0.5cm}According to  Tables \ref{IntDim} and \ref{Acc}, dimension reduction improves the accuracy and  $F_1$ scores substantially, yet  even without the dimension reduction DKPCA performs  better than the other algorithms for $5\%$ and $10\%$ for interest and line datasets and similar  for hard and serve data sets.
On the other hand, dimension reduction improves the efficiency of KNN, since KNN performs poorly for datasets with high dimensions (dimensionality curse).
  \begin{table}[!htp]
\centering
\caption{Micro $F_1$-scores on the Datasets: If the difference among two competitive scores is less than $1\%$ then we emphasized both scores.}
\label{F1-mic}
\begin{tabular}{|l|l|l|l|l|l|l|l|l|}
\hline
Dataset                   & Size \% & SVM Lin        & SVM RBF & SVM Dif.       & PCA   & KPCA Pol. & KPCA RBF & DKPCA          \\ \hline
\multirow{3}{*}{Interest} & 5\%     & 57,36          & 53,00   & 63,86          & 64,17 & 64,37     & 64,32    & \textbf{72,94} \\ \cline{2-9} 
                          & 10\%    & 65,06          & 53,15   & 65,39          & 67,41 & 66,77     & 67,65    & \textbf{77,53} \\ \cline{2-9} 
                          & 30\%    & 79,48          & 53,59   & 79,04          & 73,08 & 72,99     & 72,80    & \textbf{81,68} \\ \hline
\multirow{3}{*}{Line}     & 5\%     & 53,74          & 53,46   & 58             & 60,45 & 62,76     & 62       & \textbf{67,92} \\ \cline{2-9} 
                          & 10\%    & 56,06          & 53,45   & 61,26          & 63,88 & 66,48     & 65,57    & \textbf{71,27} \\ \cline{2-9} 
                          & 30\%    & 73,27          & 53,46   & \textbf{75,51} & 69,21 & 70,58     & 71,71    & \textbf{75,75} \\ \hline
\multirow{3}{*}{Serve}    & 5\%     & 65,27          & 41,44   & 68,71          & 63,53 & 64,04     & 62,06    & \textbf{71,73} \\ \cline{2-9} 
                          & 10\%    & 72,26          & 41,45   & 70,69          & 66,56 & 66,94     & 67,05    & \textbf{74,48} \\ \cline{2-9} 
                          & 30\%    & \textbf{81,71} & 41,48   & 78,51          & 69,81 & 70,88     & 70,47    & 77,23          \\ \hline
\multirow{3}{*}{Hard}     & 5\%     & 79,80          & 79,76   & \textbf{81,81} & 80,19 & 79,76     & 79,80    & \textbf{81,15} \\ \cline{2-9} 
                          & 10\%    & 80,12          & 79,76   & \textbf{82,54} & 80,41 & 79,86     & 79,79    & \textbf{82,31} \\ \cline{2-9} 
                          & 30\%    & 81,79          & 79,85   & \textbf{84,07} & 80,86 & 80,08     & 80       & \textbf{83,86} \\ \hline
\end{tabular}
\end{table}For hard (12055 terms), serve(17475 terms ) and line(17002 terms) datasets, we just set the dimension to be 1710 randomly, to see how random choice affects the algorithm. In  Table \ref{NoDim}  we present the scores DKPCA get when applied on these datasets without the dimension reduction. Moreover, the skewness of the distribution of the classes are significant  and this skewness  causes the micro averaged scores to be  larger than macro averaged scores as seen in Tables \ref{F1-mac} and \ref{F1-mic}.  
Table \ref{Acc}  shows that for $5\%$ and $10\%$  labeling the accuracy scores of DKPCA are significantly larger  than all the other algorithms  for the datasets line, serve and interest. This is an important result, concerning the scarcity of the labelled data. This result  also indicates that this algorithm can be converted to semi-supervised algorithm as done in \cite{Su04}. Moreover, carefully done dimension reduction on interest dataset creates a larger positive difference, circa $10\%,$  for DKPCA.    Even random or no reduction on the other sets  creates a lesser difference, circa $2\%.$
For the hard dataset  the difference between the closest Algorithm, i.e., Diffusion Kernel SVM are statistically the same  since the difference is not statistically significant. Hence,  careful parameter optimization for hard data might change the difference.  However, for $30\%$ labeled dataset,  DKPCA has a competitive performance compared to SVM Dif.  and SVM Lin., for serve dataset, SVM Lin. gives very high performance. Micro-averaged $F_1$ scores, as seen in Table \ref{F1-mic}, are very similar to accuracy scores. On the other hand, macro-averaged $F_1$ scores are very low compared to the other metrics in use. Despite this fact,  DKPCA outperforms all the other algorithms for $5\%$ and $10\%$ percent labeled data and get statistically similar results for $30\%$ labeling except for the serve dataset.

\begin{table}[!htp]
\centering
\caption{Macro $F_1$-scores on the Datasets: If the difference among two competitive scores is less than $1\%$ then we emphasized both scores.}
\label{F1-mac}
\begin{tabular}{|l|l|l|l|l|l|l|l|l|}
\hline
Dataset                   & Size \% & SVM Lin        & SVM RBF & SVM Dif.       & PCA   & KPCA Pol. & KPCA RBF & DKPCA          \\ \hline
\multirow{3}{*}{Interest} & 5\%     & 17,92          & 11,76   & 29,18          & 32,19 & 32,77     & 32,01    & \textbf{45,26} \\ \cline{2-9} 
                          & 10\%    & 26,03          & 12,32   & 34,49          & 39,31 & 38,21     & 39,97    & \textbf{52,56} \\ \cline{2-9} 
                          & 30\%    & 51,69          & 13,40   & 54,87          & 47,92 & 48,19     & 47,60    & \textbf{57,62} \\ \hline
\multirow{3}{*}{Line}     & 5\%     & 12,55          & 11,61   & 21,97          & 31,63 & 41,76     & 40,31    & \textbf{49,08} \\ \cline{2-9} 
                          & 10\%    & 19,59          & 11,61   & 28,58          & 38,88 & 47,78     & 46,73    & \textbf{54,78} \\ \cline{2-9} 
                          & 30\%    & 60,24          & 11,61   & \textbf{61,21} & 48,68 & 53,48     & 54,64    & \textbf{61,76} \\ \hline
\multirow{3}{*}{Serve}    & 5\%     & 42,59          & 14,65   & 53,43          & 47,65 & 49,66     & 46,81    & \textbf{62,94} \\ \cline{2-9} 
                          & 10\%    & 56,79          & 14,68   & 55,76          & 51,37 & 53,00     & 53,10    & \textbf{67,02} \\ \cline{2-9} 
                          & 30\%    & \textbf{75,18} & 14,73   & 70,97          & 57,77 & 59,73     & 58,80    & 70,38          \\ \hline
\multirow{3}{*}{Hard}     & 5\%     & 30,05          & 29,74   & 43,60          & 35,52 & 30,30     & 30,64    & \textbf{48,30} \\ \cline{2-9} 
                         & 10\%    & 32,63          & 29,94   & 47,19          & 36,31 & 30,79     & 30,72    & \textbf{53,77} \\ \cline{2-9} 
                          & 30\%    & 43,75          & 43,75   & 54,79          & 38,17 & 32,37     & 32.32    & \textbf{58,40} \\ \hline
\end{tabular}
\end{table}

\begin{table}[!htp]
\centering
\caption{DKPCA Scores Without Dimension Reduction for Line, Hard and Serve Datasets}
\label{NoDim}
\begin{tabular}{|l|l|l|l|l|}
\hline
Dataset                & Size & Accuracy & $F_1$-micro & $F_1$-macro \\ \hline
\multirow{3}{*}{Line}  & 5\%  & 65,17    & 65,17    & 42,95    \\ \cline{2-5} 
                       & 10\% & 68,68    & 68,68    & 48,92    \\ \cline{2-5} 
                       & 30\% & 72,77    & 72,77    & 55,79    \\ \hline
\multirow{3}{*}{Serve} & 5\%  & 67,02    & 67,02    & 53,73    \\ \cline{2-5} 
                       & 10\% & 69,56    & 69,56    & 57,93    \\ \cline{2-5} 
                       & 30\% & 72,21    & 72,21    & 62,23    \\ \hline
\multirow{3}{*}{Hard}  & 5\%  & 80,10    & 80,10    & 34,33    \\ \cline{2-5} 
                       & 10\% & 80,44    & 80,44    & 36,21    \\ \cline{2-5} 
                       & 30\% & 81,36    & 81,36    & 34,33    \\ \hline
\end{tabular}
\end{table}

\section{Conclusion and Further Study}
Experimental results show that DKPCA outperforms all kernel based SVM algorithms including the  semantic diffusion kernel based SVM for $\%5$ and $\%10$ percent train-test splits for all datasets except for hard dataset. Considering the scarcity of the labeled data, our proposed algorithm (Diffusion Kernel Principal Component Analysis) DKPCA gives promising results. 
   As stated in  \cite{Su04}  there are some disadvantages of using KPCA based algorithms for datasets that contain targets with dissimilar contexts. In the same paper,  this issue is resolved by converting the supervised method to semi-supervised method by including unlabeled data for creating the kernel. As future work, we want to apply this semi-supervised technique with DKPCA  and compare our results.  
   
Instead of employing  DKPCA only for WSD problems, it can be applied to other research areas in machine learning, i.e.,  DKPCA can be quite versatile. For example,  DKPCA can be used for feature extraction and dimension reduction based on higher order correlation of the data, and this approach may lead to hybrid DKPCA models with  other machine learning algorithms or neural networks. We believe that it is a good starting point for merging DKPCA with other  deep learning methods.  We expect that DKPCA boots the other algorithms performance  when there are higher order similarities among data \cite{Ganiz06}.

%
%

\begin{thebibliography}{8}

\bibitem{Altinel15}
Altınel, B., Ganiz, M.C., Diri, B.:  A corpus-based semantic kernel for text classification by using meaning values of terms. Eng. Appl. Artif. Intell. 43, 54–66.(2015)

\bibitem{Bruce94}
 Bruce, R.F., Wiebe, J.: Word-sense disambiguation using decomposable models. 
 In: Proceedings of the 32nd Annual Meeting of the Association for Computational Linguistics, pp. 139–146 ACL Las Cruces, USA, (1994).
 
\bibitem{Carpuat07}
Carpuat M, Wu, D.:  Improving statistical machine translation using word sense disambiguation. In: Proceedings of the 2007 joint conference on empirical methods in natural language processing and computational natural language learning, pp 61–72 Czech Republic,  Prague,  (2007)

\bibitem{Cortes95}
Cortes, C., Vapnik, V. :Support-vector networks. Mach Learn 20(3):273–297 (1995) 


\bibitem{Cristianini02}
Cristianini, N., Shawe-Taylor, J., Lodhi, H.:Latent semantic kernels. J. Intell. Inf.
Syst. 18 (2–3), 127–152. (2002)


\bibitem{Ganiz06}
Ganiz, M. C.,  Pottenger W. M. and Lehigh X. Y. :Link Analysis of Higher-Order Paths in Supervised Learning Datasets, In Proceedings of the Workshop on Link Analysis, Counterterrorism and Security, SIAM Conference on Data Mining, Bethesda, MD, April (2006) 

\bibitem{Gliozzo05}
Gliozzo, A., Giuliano, C.: Strapparava, C.: Domain kernels for word sense disambiguation. In: Proceedings of the 43rd Annual Meeting of the Association for
Computational Linguistics,  pp. 403–410. University of Michigan, USA (2005)

\bibitem{Gonen11}
Gönen M, Alpayın E :Multiple kernel learning algorithms. J Mach Learn Res  12:2211–2268 (2011)


\bibitem{Hotelling33}
 Hotelling, H.: Analysis of a Complex of Statistical Variables Into Principal Components. Warwick and York, (1933)

\bibitem{Hofman08}
Hofmann, T.,   Schölkopf, B., Smola, A.J.:Kernel methods in machine learning.Annals of Statistics, 36(3):1171–1220, (2008)

\bibitem{Iacobacci16}
Iacobacci, I,   Pilehvar , M.T., Navigli, R. :Embeddings for Word Sense Disambiguation: An Evaluation Study.  In Proc. of ACL, pp. 897-907, (2016)

\bibitem{Jurafsky10}
Jurafsky, D.- Martin, J. H. :Speech and language processing (Vol. 3),  Pearson, London  (2014)

\bibitem{Kandola03}
Kandola, J., Shawe-Taylor, J., Cristianini, N. : Learning semantic similarity. Adv Neural Inf Process Syst 15:657–664 (2003)

\bibitem{Kondor02}
Kondor, R. I.   and  Lafferty J.: Diffusion kernels on graphs and other discrete input spaces. In Proceedings of the ICML (2002)

\bibitem{Kondor04}
Kondor, R. I., Vert,  J.P. :Diffusion kernels. In: Schölkopf B, Tsuda K, Vert J P, editors. Kernel methods in computational biology. Cambridge (Massachusetts): MIT Press.  171--192.  (2004)

\bibitem{Leacock93}
Leacock, C., Towell, G.,Voorhees, E. :Corpus-based statistical sense resolution. In: Proceedings of the ARPA Workshop on Human Language Technology, pp. 260–265. Plainsborogh, USA (1993)

\bibitem{Leacock98} 
Leacock, C., Miller, G.A, Chodorow, M. :Using corpus statistics and WordNet relations for sense identification Comput. Linguist., 24 (1) , pp. 147-165 (1998)

 \bibitem{Li16} 
 Li,X., Qing, S., Zhang, H., Wang, T., Yang, H.: Kernel methods for word sense disambiguation Artif. Intell. Rev., 46 (1) , pp. 41-58 (2016)


\bibitem{McCaughren09}
McCaughren A.:Polysemy and Homonymy and their Importance for the Study of Word Meaning, ITB Journal Published (2009)
 \doi{10.21427/D7SJ17}

\bibitem{Muller01}
M$\ddot{u}$ller K.R., Mika S., Rätsch, G., Tsuda, K., Schölkopf, B.  :An introduction to kernel-based learning algorithms. IEEE Trans Neural Netw 12(2):181–202 (2001)

\bibitem{Navigli09}
Navigli R: Word sense disambiguation: a survey. ACM Comput Surv   41(2):1–69 (2009)


\bibitem{Nguyen13}
Nguyen KH, Ock C.Y.:Word sense disambiguation as a traveling salesman problem.   Artif Intell Rev 40(4):405–427 (2013)

\bibitem{Pearson01}
K. Pearson. :On lines and planes of closest fit to systems of points in space. Philosophical
Magazine Series 6, 2(11):559–572, (1901)

\bibitem{Scholkopf02} 
Schölkopf, B.-Smola, A.:Learning with Kernels. MIT Press, Cam-bridge, MA.(2002).

\bibitem{Scholkopf98} 
Schölkopf, B. , Smola, A. and   Müller, K : Nonlinear component analysis as a kernel eigenvalue problem. Neural Computation,10(5), (1998).

\bibitem{Shawe-Taylor04}
Shawe-Taylor J, Cristianini N.: Kernel methods for pattern analysis. Cambridge University Press, New York (2004)


\bibitem{Shawe-Taylor02}
Cristianini, N., Shawe-Taylor, J., Lodhi, H.: Latent semantic kernels. J. Intell. Inf.
Syst. 18 (2–3), 127–152.(2002)

\bibitem{Su04}
Su W, Carpuat M, Wu, D.: Semi-supervised training of a kernel PCA-based model for word sense disambiguation. In: Proceedings of the 20th international conference on computational linguistics,  pp 1298–1304, Geneva (2004)

\bibitem{Turdakov10}
Turdakov, D.Y.:Word sense disambiguation methods. Program Comput Softw    36(6):309–326 (2010)

\bibitem{Vidal16}
Vidal, R., Ma, Y., Shankar, S.: Generalized Principal Component Analysis,Interdisciplinary Applied Mathematics 40 Springer (2016)


\bibitem{Wang13}
Wang, T., Rao, J., Zhao, D.: Using exponential kernel for word sense disambiguation. In: Proceedings of the 23rd international conference on artificial neural networks, pp 545–552 LNCS 8131, Sofia, (2013)

\bibitem{Wang14}
 Wang, T.,   Rao J.,  Hua Q.:Supervised word sense disambiguation using semantic diffusion kernel. Engineering Applications of Artificial Intelligence, 27. Elsevier, 167--174,  (2014)
 
\bibitem{Wang17}
 Wang, T., Li W., Liu, F. and  Hua J.: “Sprinkled semantic diffusion kerne(l for word sense disambiguation”, Engineering Applications of Artificial Intelligence., 64, no. May,  43--51, (2017)
 
 \bibitem{Wu04}
 Wu, D., Su, W. and Carpuat M.:  A Kernel PCA method for superior word sense disambiguation.  In Proceedings of the 
42nd Annual Meeting of the Association for Computational Linguistics, Barcelona, Spain, July (2004)



\end{thebibliography}
%

\end{document}